\documentclass[twoside,11pt]{article}
\usepackage{jmlr2e}
\usepackage{amsmath}

\jmlrheading{2022}{cs.LG}{1-7}{12/22}{12/22}{}{Dumitru Damian}
\ShortHeadings{Lightmorphic Signatures Analysis Toolkit}{Damian}
\firstpageno{1}
\begin{document}
\title{Lightmorphic Signatures Analysis Toolkit}
\author{\name Dumitru Damian \email dumitrudamian@yahoo.com \\
       \addr Information and Communication Engineering\\
       \addr Research and development consultant\\
       \addr Timișoara, RO}
\maketitle
\begin{abstract}%
In this paper we discuss the theory used in the design of an open source lightmorphic signatures analysis toolkit (LSAT). In addition to providing a core functionality, the software package enables specific optimizations with its modular and customizable design.

To promote its usage and inspire future contributions, LSAT is publicly available. By using a self-supervised neural network and augmented machine learning algorithms, LSAT provides an easy-to-use interface with ample documentation. 

The experiments demonstrate that LSAT improves the otherwise tedious and error-prone tasks of translating lightmorphic associated data into usable spectrograms, enhanced with parameter tuning and performance analysis.  

With the provided mathematical functions, LSAT validates the nonlinearity encountered in the data conversion process while ensuring suitability of the forecasting algorithms.
\end{abstract}

\section{Introduction} 
It is common knowledge, in the machine learning domain, to use differential values, since they provide a simple way to model the data. However, such algorithms may not fit the lightmorphic signature properly, leading to a reduced quality of the obtained results. Training a neural network to predict the lightmorphic signature can significantly increase the data quality. This is the task that LSAT tries to accomplish.

As such we define the lightmorphic metric learning (LML) as a branch of machine learning algorithms, set out with the purpose of learning lightmorphic signatures from multiple datasets trough usage of vibrating graph chords.

In the pursuing sections we describe the main features of the toolkit, explain the general mathematical concepts and finally detail the plans regarding future functionalities.
\section{General mathematical concepts}
In this section we expand the mathematical concepts and link them with the reasoning encountered in the implemented code. 

We define the lightmorphic signature as a function of: light intensity (I) that varies according to seasons and local weather conditions, trajectory distribution characteristics (D), and specific adjustments (T):
\begin{equation}  \label{eq:light signature function}
f_{L_{\odot}} =  \int\limits_{1}^{I}\int\limits_{1}^{D}\int\limits_{1}^{T} \Gamma_{t}\zeta_{t}{dt}
\end{equation}
where:
\begin{itemize}
	\item $\Gamma_{t}$ -- trajectory tensor
	\item $\zeta_{t}$ -- point in time specificity
\end{itemize}

Storage of these trajectory specific lightmorphic signatures is done in a database ($\Theta$). The segments containing isochronous surfaces with similarities are stored in another database ($\Phi$) that serves as a baseline for training the neural network implementation. 

The isochronous surfaces that constitute the lightmorphic signature are interlinked trough the definition and usage of graph chords ($\delta(t)$). Observing their vibrational amplitude allows the prediction of alternative lightmorphic signatures and, at the same time, correction of the already known values. 

Since the primary light source considered is the Earth's Sun, specific spacetime metrics (ex. $g_{\mu\nu}, \eta_{\mu\nu}, h_+, h_\times, G_{\mu\nu}$) have to be used in order to describe the encountered anisotropies. These are implemented as a function of distant astrophysical forces that stretch and compress the fabric of spacetime. 

According to special relativity, spacetime is seen as a four dimensional manifold described by a flat Minkowski metric defined in Cartesian coordinates (t, x, y, z, c = 1) as:

\begin{equation} \label{Minkowski}
\eta_{\mu\nu}= 
\begin{pmatrix} 
-1 & 0 & 0 & 0 \\
0 & 1 & 0 & 0 \\
0 & 0 & 1 & 0 \\
0 & 0 & 0 & 1 \\
\end{pmatrix} ,
\end{equation}

When considering the geometry of curved space, we have made use of the metric $g_{\mu\nu}$, that replaces the flat Minkowski metric $\eta_{\mu\nu}$. This substitution was done considering that the geometry of curved space will eventually reduce to the flat spacetime of special relativity at a sufficiently small scale.

The interaction between curvature of spacetime and the mass distribution was modeled following (\cite{Blackburn}) work, as:
\begin{equation} \label{spacetime curvature}
G_{\mu\nu} = k T_{\mu\nu}
\end{equation} where $G_{\mu\nu}$ is defined as the Einstein curvature tensor, $T_{\mu\nu}$ is the stress-energy tensor and represents the mass-energy distribution, while k describes the Einstein constant of gravitation defined as:
\begin{equation} \label{Einstein_equation_1}
k = \frac{8\pi G}{c^4}
\end{equation} where c is the speed of light in a vacuum.

At the same time, in order to improve the results quality, the Einstein tensor was also considered under the form:
\begin{equation} \label{Einstein_equation_2}
G_{\mu\nu} = R_{\mu\nu}-\frac{1}{2}g_{\mu\nu}R,
\end{equation}
where $R_{\mu\nu}$ is the Riemann tensor for the local spacetime, and R is the Ricci scalar.

Since there is not one general solution for the complex Einstein equations, but a large variety of possible solutions that apply to particular circumstances, we've considered a weak-field approximation, where the nonlinear Einstein equations where approximated towards linearity. 

For example, a very small perturbation specific to a gravitational wave, will impact the flat spacetime and it is defined as $h_{\mu\nu}(x)$ and it's value will be $|h_{\mu\nu}| << 1$. 

Thus, the Einstein equation becomes:
\begin{equation}  \label{small_disturbance}
g_{\mu\nu}(x) = \eta_{\mu\nu} + h_{\mu\nu}(x) . 
\end{equation} 

or by simply considering the induced strain variations:
\begin{equation}  \label{Einstein_Alembertian}
\Box h_{\mu\nu}(x)=0,
\end{equation}

By further pursuing such linearization, we can represent in the TT gauge, a propagating wave, under the following form: 
\begin{equation} \label{TT gauge}
h_{\mu\nu}^{TT}= 
\begin{pmatrix} 
0 & 0 & 0 & 0 \\
0 & h_{+} & h_{\times} & 0 \\
0 & h_{\times} & -h_{+} & 0 \\
0 & 0 & 0 & 0 \\
\end{pmatrix} ,
\end{equation}

where the constant amplitudes ($h_+$, $h_\times$) represent the two gravitational wave polarizations, the plus- and cross-polarization.

We represent the distance between two neighboring points as defined by (\cite{Behnke}) for a flat spacetime, trough the following expression:

\begin{center}
$ds^2 = -c^2dt^2+dx^2+dy^2+dz^2=-c^2dt^2+[1+h_+(t)]dx^2+[1-h_+(t)]dy^2 $
\end{center}

That allows us to model in the TT gauge, the gravitational wave stretching along the x axis and compression along the y axis with the specific factor of: $\sqrt{1\pm h_+(t)}\simeq 1+\frac{1}{2}h_+(t)$

Having modeled the photon's traveling path in outer space, in order to simplify the inherent path inhomogeneities, we separated the domains into outer space domain, atmospheric domain and Earth specific domains (lithosphere, hydrosphere, biosphere, noises, etc). 

We further define the phase of an electromagnetic wave of frequency $\omega_0$ as $\phi$. Following \cite{Driggers}'s work, we consider that the starting light phase is at 0 and it travels at the speed of light c. After a distance L it will have a phase $\delta_{\phi_{space}}$ that can be expressed as a distance integral over the spacetime metric,
\begin{equation} \label{space domain}
\delta_{\phi_{space}} = \frac{\omega_0}{c} \int_{0}^{L}g dx, 
\end{equation}
with $g(t)=\eta+h(t)$, where $\eta$ is the Minkowski metric and h(t) is the dimensionless spacetime strain.

Summing the light phase shift $\delta_{\phi_{atm}}$ and the $\delta_{\phi_{Earth}}$ which is derived from the noise sources like seismic or electromagnetic interferences, leads to the dataset of trajectory specific lightmorphic signatures:
\begin{equation}  \label{eq:light intensity trajectories similarities}
\Phi_{\Gamma_{IDT}} = \sum_{j=1}^{\mathbb{N}}{\Gamma_{IDT}^j}
\end{equation}

The signature parameter estimation is performed considering a prior distribution p($\Phi|L_{\odot}$) that is updated upon receiving the new data d to give a posterior distribution p($\Phi|d, L_{\odot}$)
\begin{equation}  \label{distribution}
p(\Phi|d, L_{\odot}) = \frac{p(\Phi|L_{\odot}) p(d|\Phi, L_{\odot})}{p(d|L_{\odot})}
\end{equation}

While observing the distribution of multiple light segments within the dataset $\Phi_{\Gamma_{IDT}}$, it will be possible to estimate the probability for trajectory specific lightmorphic evolution: 

\begin{equation}  \label{eq:light intensity trajectories probability}
p_{\Phi} =  f({\rho_{k} \cdot p_{\Phi_{k}}})
\end{equation}

where $p_{\Phi_{k}}$ is the database's k-th segment specific probability, $\rho_{k}$ is the prediction weight for the k-th segment. 
\section{Software package design}
The distribution matrices specific to the isochronous segmentation surfaces, which define the lightmorphic signature model, form the LSAT core. 

As such we've used a design principle that ensures simplicity for the whole package, while making the source codes easy to read and maintain. As the toolkit is written in a modular way, new functionalities can be easily plugged in. This makes the LSAT not only a lightmorphic signature machine learning tool but also an experimental platform.

LSAT comes with plenty of documentation for all the interface functionalities and related data structures. The README file describes the installation process and interface usage. For developers who use the toolkit in their applications, the API documentation can provide additional information related to functionality calls. 
\section{Practical Usage}
In the examples, we provide sample values for the lightmorphic signature updates, as a function of $\delta_{\phi_{atm}}$ derived by the neural network from the values of a large dataset of atmospheric meteorological data for 317 cities in Romania, with hundreds of thousands data points. Automatic learning is supported trough API calls to the domain specific data providers.

Beyond this simple way of running the lightmorphic signatures analysis toolkit, there are several enhancement options for advanced usage. As example, one may activate additional functionalities that consider input parameters like complex space weather forecasting, different electromagnetic wave disturbances or lithosphere, hydrosphere and biosphere specific localized data.
\section{Conclusion and Future Work}
With the lightmorphic signatures analysis toolkit we provided an open source SW package that is simple and easy-to-use. 

Experiments and analysis conclude that the modular design and customization support are performing excellent in practice and can provide the base for additional research on lightmorphic signatures. 

The toolkit is constantly being improved by new research results and user feed-back with the ultimate goal of having an automated toolkit to use in maintaining and updating a large database of high-quality light signatures. 

Future work will focus on probability estimates, additional functionalities that mitigate the large uncertainties in the available observational input data which arise from the complex interaction processes. In addition, the inclusion of artificial intelligence (AI) options will be considered while building a national/international network for lightmorphic signature analysis.
\newpage
\section{References}
\nocite{*}
\bibliography{Lightmorphic_Signatures_Analysis_Toolkit.bib}

\begin{thebibliography}{16}
\providecommand{\natexlab}[1]{#1}
\providecommand{\url}[1]{\texttt{#1}}
\expandafter\ifx\csname urlstyle\endcsname\relax
  \providecommand{\doi}[1]{doi: #1}\else
  \providecommand{\doi}{doi: \begingroup \urlstyle{rm}\Url}\fi

\bibitem[Adhikari(2004)]{Rana}
Rana Adhikari.
\newblock \emph{{Sensitivity and noise analysis of 4 km laser interferometric
  gravitational wave antennae}}.
\newblock PhD thesis, Massachusetts Institute of Technology, 2004.

\bibitem[Berit(2013)]{Behnke}
Behnke Berit.
\newblock \emph{{A Directed Search for Continuous Gravitational Waves from
  Unknown Isolated Neutron Stars at the Galactic Center}}.
\newblock PhD thesis, Leibniz University Hannover, 2013.

\bibitem[Biscoveanu et~al.(2021)Biscoveanu, Isi, Vitale, and
  Varma]{Biscoveanu:2020are}
Sylvia Biscoveanu, Maximiliano Isi, Salvatore Vitale, and Vijay Varma.
\newblock {New Spin on LIGO-Virgo Binary Black Holes}.
\newblock \emph{Phys. Rev. Lett.}, 126\penalty0 (17):\penalty0 171103, 2021.
\newblock \doi{10.1103/PhysRevLett.126.171103}.

\bibitem[Blackburn(2010)]{Blackburn}
Lindy Blackburn.
\newblock \emph{{Open Issues in the Search for Gravitational Wave Transients}}.
\newblock PhD thesis, Massachusetts Institute of Technology, 2010.

\bibitem[Clark(2013)]{Clark}
Daniel~E. Clark.
\newblock \emph{{Control of Differential Motion between Adjacent Advanced LIGO
  Seismic Isolation Platforms}}.
\newblock PhD thesis, Stanford University, 2013.

\bibitem[Dooley(2011)]{Dooley}
Katherine~Laird Dooley.
\newblock \emph{{Design and performance of high laser power interferometers for
  gravitational-wave detection}}.
\newblock PhD thesis, University of Florida, 2011.

\bibitem[Driggers(2015)]{Driggers}
Jennifer~Clair Driggers.
\newblock \emph{{Noise Cancellation for Gravitational Wave Detectors}}.
\newblock PhD thesis, California Institute of Technology, 2015.

\bibitem[Fricke(2011)]{Fricke}
Tobin~Thomas Fricke.
\newblock \emph{{Homodyne detection for laser-interferometric gravitational
  wave detectors}}.
\newblock PhD thesis, Louisiana State University and Agricultural and
  Mechanical College, 2011.

\bibitem[Lasky et~al.(2016)]{Lasky:2015lej}
Paul~D. Lasky et~al.
\newblock {Gravitational-wave cosmology across 29 decades in frequency}.
\newblock \emph{Phys. Rev. X}, 6\penalty0 (1):\penalty0 011035, 2016.
\newblock \doi{10.1103/PhysRevX.6.011035}.

\bibitem[Maggiore(2000)]{Maggiore:1999vm}
Michele Maggiore.
\newblock {Gravitational wave experiments and early universe cosmology}.
\newblock \emph{Phys. Rept.}, 331:\penalty0 283--367, 2000.
\newblock \doi{10.1016/S0370-1573(99)00102-7}.

\bibitem[Martynov(2015)]{Martynov}
Denis Martynov.
\newblock \emph{{Lock Acquisition and Sensitivity Analysis of Advanced LIGO
  Interferometers}}.
\newblock PhD thesis, California Institute of Technology, 2015.

\bibitem[Quitzow-James(2016)]{Ryan}
Ryan Quitzow-James.
\newblock \emph{{Search for Long-Duration Transient Gravitational Waves
  Associated with Magnetar Bursts during LIGO\textquoteright{}s Sixth Science
  Run}}.
\newblock PhD thesis, Oregon U., 2016.

\bibitem[Romano and Cornish(2017)]{Romano:2016dpx}
Joseph~D. Romano and Neil~J. Cornish.
\newblock {Detection methods for stochastic gravitational-wave backgrounds: a
  unified treatment}.
\newblock \emph{Living Rev. Rel.}, 20\penalty0 (1):\penalty0 2, 2017.
\newblock \doi{10.1007/s41114-017-0004-1}.

\bibitem[Ross(2020)]{Ross}
Michael~P. Ross.
\newblock \emph{{Precision Mechanical Rotation Sensors for Terrestrial
  Gravitational Wave Observatories}}.
\newblock PhD thesis, University of Washington, 2020.

\bibitem[Tuyenbayev(2017)]{Tuyenbayev}
Darkhan Tuyenbayev.
\newblock \emph{{Extending the scientific reach of Advanced LIGO by
  compensating for temporal variations in the calibration of the detectors}}.
\newblock PhD thesis, The University of Texas at San Antonio, 2017.

\bibitem[Wade(2015)]{Wade}
Madeline Wade.
\newblock \emph{{Gravitational-Wave Science with the Laser Interferometer
  Gravitational-Wave Observatory}}.
\newblock PhD thesis, University of Wisconsin–Milwaukee, 2015.

\end{thebibliography}
\end{document}